# Robustness Evaluation for Video Models with Reinforcement Learning

Ashwin Ramesh Babu,* Sajad Mousavi*, Vineet Gundecha, Sahand Ghorbanpour,
Avisek Naug, Antonio Guillen, Ricardo Luna Gutierrez, Soumyendu Sarkar[†]
Hewlett Packard Enterprise (Hewlett Packard Labs)
{ashwin.ramesh-babu, sajad.mousavi, vineet.gundecha
sahand.ghorbanpour, avisek.naug, antonio.guillen, rluna
soumyendu.sarkar}@hpe.com

## Abstract

*Evaluating the robustness of Video classification models is very challenging, specifically when compared to image-based models. With their increased temporal dimension, there is a significant increase in complexity and computational cost. One of the key challenges is to keep the perturbations to a minimum to induce misclassification. In this work, we propose a multi-agent reinforcement learning approach (spatial and temporal) that cooperatively learns to identify the given video's sensitive spatial and temporal region. The agents consider temporal coherence in generating fine perturbations leading to a more effective and visually imperceptible attack. Our method outperforms the state-of-the-art solutions on $L_p$ metric and the average queries. Our method enables custom distortion types, making the robustness evaluation more relevant to the use case. We extensively evaluate 4 popular models for video action recognition on two popular datasets HMDB-51 and UCF-101.*

## 1. Introduction

Convolutional neural networks (CNNs) excel in various computer vision tasks such as image classification, object detection, tracking, and video recognition. Despite their success, CNNs are vulnerable to adversarial examples—small, deliberate perturbations that mislead the models. These adversarial examples can be generated via white-box attacks, which require full access to the model's parameters, or black-box attacks, which do not require in-depth knowledge of the model but are harder to execute, requiring more queries and distortions. Black-box attacks often use query-based methods, where the attacker iteratively queries the victim model to gather prediction scores and refine subsequent queries. Videos add complexity with temporal information, increasing computational cost, query numbers, and perturbations. Therefore, reducing the spatial and temporal search space is essential for generating successful adversarial samples with fewer queries.

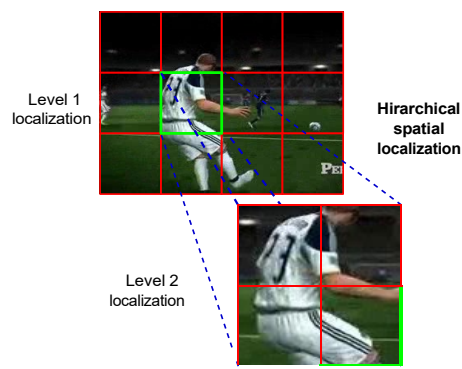

Figure 1. The designed actions of the spatial agent. For each frame, the space is first divided into a set of non-overlapping grids from which the agent localizes the action region, and the chosen patch is further divided into non-overlapping patches from which the final action is decided by the spatial agent. This reduces the search space significantly and the $L_p$ norm.

In this work, inspired by the work of [18, 21], we introduce a novel multi-agent reinforcement learning (RL) setting that has a spatial agent that learns to identify the most sensitive location in a given frame and a temporal agent that learns to recognize the most impactful frames in a given video sample. These two agents are jointly trained by a common objective which is to induce misclassification by the victim model. The agents train in an iterative approach to identify the sensitive space composed of spatial and temporal regions while adding distortion. The temporal agent consists of long-short-term memory network (LSTM) blocks to capture the relation between the consecutive and successive frames, deciding to include or not include the current frame. The spatial agent adopts a hierarchical approach where the

---

*Equal Contribution
[†]Corresponding Author

region of interest is identified in two steps as shown in Figure 1. First, a larger localization (L1 localization) of the event happening in the video is identified by dividing the input frames in a video into a set of non-overlapping grids and the agent takes an action by choosing one of the patches in the video frame. Further, to refine the focus region, the output of the L1 localization is divided into a non-overlapping grid and the agent chooses the second action narrowing it to a finer patch to which perturbation is added. Both agents have a common reward and task-specific rewards to introduce a successful attack on the victim models. The hierarchical approach was adopted for spatial agents to significantly reduce the search space and for faster convergence. Finally, a distortion reversion is performed to remove distortions that were previously considered important but do not have any significant impact at the final state. This way, the minimum $L_p$ norm is maintained. The main contribution of the paper can be summarized as,

- A novel multi-agent reinforcement learning-based black-box adversarial attack method that aims at evaluating the robustness of video recognition models.
- The RL agents are composed of a spatial agent that aims at identifying the most sensitive spatial region in a **hirarchical approach**. The temporal agent is responsible for identifying image frames that are highly impactful in a given input video.
- A reverse approach that aims at removing distortions that are less impactful at a given state to cause misclassification
- When comparing the results with the current state-of-the-art approaches on video attack, we perform significantly better in maintaining low $l_p$ norm and query numbers.

## 2. Related Works

### 2.1. Query-based Attacks

Recent research on black-box video attacks has focused on transfer attacks and query attacks, each with pros and cons. Transfer attacks leverage the transferability of adversarial inputs to videos, transferring the attack across different target videos and victim models [13]. Notable efforts include [1, 10, 31, 32]. Query-based methods, while more widely adopted, can be costly in terms of the number of queries needed due to the high dimensionality of video data. Several works have aimed to improve these methods by reducing data dimensions but still face the challenge of high query numbers [8, 11, 30, 36].

[11] proposed an iterative algorithm using geometric transformations to reduce the search space, improving gradient estimation for misclassification. [6] found that action recognition models are vulnerable to attacks on a single video frame. [8] showed that using Image-Net pre-trained models for perturbations can reduce queries for video attacks, suggesting similarity in intermediate features between image and video models. Extending this, [33] designed a cross-modal attack generating adversarial frames to attack video models. Conversely, [36] argued against using random noise and incorporated motion information in generating adversarial examples. [35] proposed a cube-based tiling strategy and a random-search evolution algorithm to reduce data dimensions and search space. [27] used $l_{2,1}$ norm-based optimization to propagate perturbations across frames, reducing the need for additional perturbations.

### 2.2. Reinforcement Learning Based Attacks

Some works have utilized reinforcement learning (RL) to find sparse key frames or important patches within frames to improve query efficiency [26, 29, 34]. [34] and [26] used RL for frame selection, with the former's agent updated only after successful attacks, leading to unnecessary queries, while the latter used saliency maps for key region selection. [29] proposed a multi-agent RL setup with agents selecting frames and patches, sharing a common reward, and used Projected Gradient Descent (PGD) with Natural Evolution Strategy (NES) for the attack.

[7] introduced a tiling strategy based on the idea that nearby pixels tend to be similar, estimating gradients for patches rather than individual pixels for higher efficiency. [14–20, 24] approaches applied a similar patch-based approach, using a tree search to add noise non-exhaustively, achieving efficient results. In this proposed work, we use a multi-agent reinforcement learning strategy where one agent optimizes to choose the spatial region in the image frames, and the other aims at selecting the right set of frames. Our work introduces a novel hierarchical strategy to reduce the search space and introduces a reversing strategy to keep the perturbation to a minimum.

## 3. Background

**Reinforcement Learning (RL):** In this work, we model reinforcement learning using a Markov Decision Process framework, denoted by $< S, A, R, P, \gamma >$, where $S$ is the state space, $A$ is the action space, $R: S \times A \to \mathbb{R}$ is the reward function, and $P: S \times A \times S \to [0, 1]$ is the transition kernel which denotes the probability of transitioning to a state $s'$ while taking an action $a$ in state $s$ and $\gamma$ denotes the discount factor. A policy $\pi: S \times A \to [0, 1]$ is a function that determines the probability of taking action $a$ in state $s$. The goal of RL is to find a policy that maximizes the infinite horizon discounted reward $J(\pi) = \mathbb{E}_{\pi, P, \mu_0} [\sum_{t=0}^{\infty} \gamma^t R(s_t, a_t)]$, where $\mu_0$ is the initial state distribution. In this work we use PPO [22], a popular actor-ciritc algorithm to optimize the RL objective.

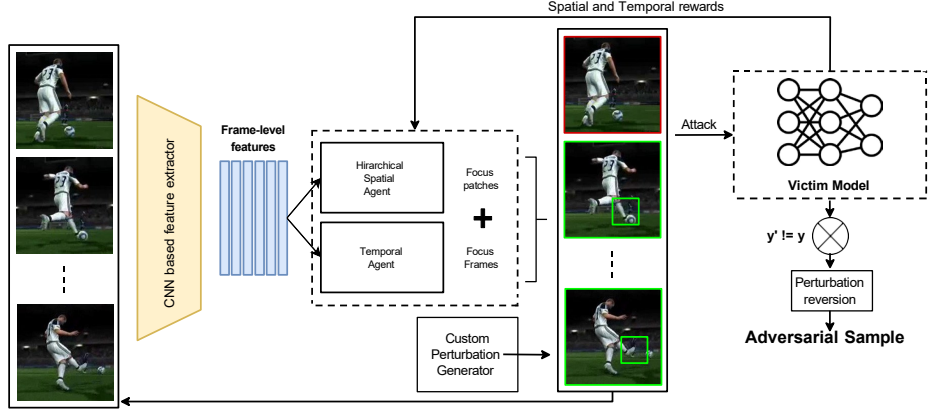

Figure 2. Overview of the proposed method. A CNN-based architecture is used to extract features from the video frames, which are in turn passed to the temporal agent and the hierarchical spatial agent to decide on the focus frames and the focus patches to which perturbations are added, and the victim model is attacked until a misclassification happens.

## 4. Problem Setup

We have a victim model $V(x) = y$ that maps a video $x$ to its corresponding label $y$, where $x, y \sim D$ are drawn from our database D. Our goal is to add distortions to $x$ to obtain $x'$, such that the victim model misclassifies it $V(x') \neq y$. We do this iteratively, starting with a clean video $x_0$. The agents add distortions to this video to generate $x_1$, receives a reward, and the process continues. The video $x_t = [f_t^1, f_t^2, \ldots, f_t^M]$ at each iteration $t$ consists of $M$ frames, where $f_t^i$ denotes the $i$th frame at iteration $t$.

## 5. Proposed Method

The proposed method uses a multi-agent reinforcement learning strategy with two agents: one identifies "focus frames" in a video, and the other identifies the "focus region" within each selected frame. The overall architecture is shown in Figure 2. These agents work together towards a common goal, formalized as a Markov Decision Process. To simplify, the magnitude of the added noise remains constant, regardless of the noise type. Detailed specifics of the two agents and the process are provided in the following sections.

### 5.1. Temporal Agent

The temporal agent aims at mining the most impactful frames for a given video. The design of the temporal agent was inspired by the work of [3]. For every image frame considered, the agent acts as a binary classifier where 0 signifies that the given frame is excluded at a given point and 1 represents the frame to be considered for perturbation in a given time step. The total number of frames chosen by the agent will be less than the total frames in the given input video.

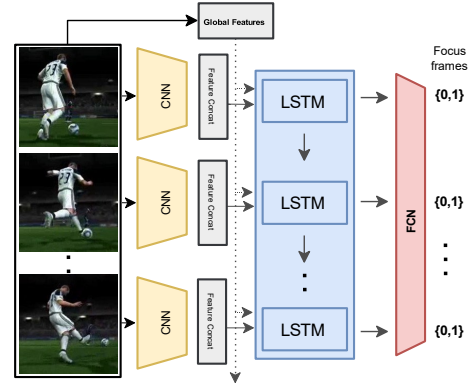

Figure 3. Temporal agent

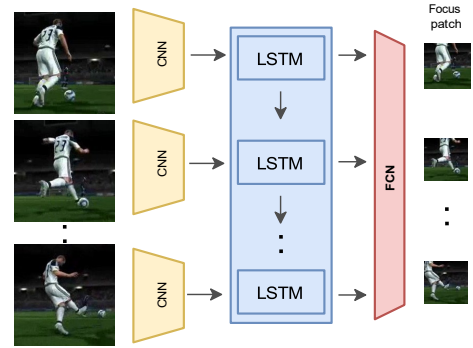

Figure 4. Spatial Agent

Figure 5. Policy network architecture of the spatial (bottom) and temporal (top) agent

#### 5.1.1. Policy Network

The temporal policy network $\pi_{\theta_C}$, parameterized by $\theta_C$ takes an action $a_t^i$ to decide if frame $i$ is selected ($a_t^i = 1$) or discarded ($q^i = 0$) at iteration $t$. The input to the policy network are the the current frame $f_t^i$, the global features of the video $g_t$, and information from the past frames $h_t^{i-1}$.

$$a_t^i \sim \pi_{\theta_c}(\cdot | f_t^i, g_t, h_t^{i-1}) \quad i = 1, 2, \ldots, M$$

The action is computed for all the frames for video $x_t$. The policy network is composed of a feature extractor (i.e., CNNs) and a frame-level classifier. For the feature extractor, we use a pre-trained MobileNet-V2 model. The agent is an LSTM topped with a fully connected (FC) layer as shown in Figure 3.

### 5.1.2. Reward

The reward for the temporal agent is a combination of three components consisting of a common reward shared with the spatial agent, and is provided after all the frames have been parsed. The first reward defined as $r_t^1$ aims at optimizing the agent in choosing the minimum number of frames for a given video. This is achieved by setting a threshold hyperparameter $L$ which is less than the total number of frames $M$. Hence the reward can be defined as,

$$r_t^1 = \exp\left(-\frac{1}{M}\left|\sum_{i=1}^{M} a_t^i - L\right|\right) \quad (1)$$

Inspired by [37], the second reward function aims at choosing the subset of frames that best represents the whole video sequence as the subset selected should be sparse while semantically representative of the whole video. Hence, $r_t^2$ can be represented as,

$$r_t^2 = \exp\left(-\frac{1}{M}\sum_{i=1}^{M} \min_{i \in K_t} \|e_t^i - e_t'\|_2\right) \quad (2)$$

where $K_t$ represents the set of the selected frame and $e_t^i$ represents frame-level feature embedding.

The common reward $r_t^3$ is the final feedback from the victim models. It measures the impact of the added noise on a selected patch. If the selected patch is reasonable, the generated adversarial patch should have a strong impact on decreasing the confidence score of the initial prediction. This is represented as,

$$r_t^3 = \exp(L(V(x_{t+1}) - V(x_t))) \quad (3)$$

where $x_t$ represents the video at the beginning of the iteration and $x_{t+1}$ represents the perturbed video at the end of iteration $t$. $L$ is a loss function that computes the difference in confidence scores between $x_t$ and $x_{t+1}$.

### 5.2. Spatial Agent

The spatial agent has two levels to localize the most impactful region to which perturbations will be added. The first level narrows down to a larger action patch in a given frame followed by a second level which further refines the region of interest. To reduce the complexity of the problem, the selection of spatial patches was divided into two distinct processes.

### 5.2.1. Level-1 localization

The action space for the level-1 localization divides each video frame into large non-overlapping patches, this similar to the method proposed in [18]. For a given image frame $f_t^i$ from a video, the level-1 action for the spatial agent $b_t^{i,1}$ ranges from $[1, D]$ where $D$ represents the total number of patches for a given frame. The spatial agent's goal is to choose the right patch for each frame in the video. Hence the final action is a sequence represented as

$$A_p = \{a_t^i | i = 1....M\} \quad (4)$$

where $M$ represents the total number of video frames and $p$ represents the spatial action list.

### 5.2.2. Policy Network

The spatial policy network, denoted as $\pi_p(A_p|s_p)$, predicts spatial actions $A_p$ based on the state $s_p$. As shown in Figure 4, an LSTM-based architecture is used to build the policy network. Initial features are extracted from image frames using a pre-trained MobileNet-V2 CNN backbone with frozen weights. These features are then fed into an LSTM followed by a fully connected layer to map the features to the output layer. The LSTM architecture ensures that regions identified in consecutive frames account for previous predictions, maintaining effective object localization. The agent's states consist of features from both current and preceding frames. The policy network is updated iteratively to enhance the optimal action region until misclassification occurs.

### 5.2.3. Reward

Similar to the temporal agent, the spatial agent's reward structure consists of three components. In each iteration, the spatial policy network receives feedback from the environment to adjust its parameters $\theta_p$. Designing suitable rewards is crucial for guiding the policy network. For level-1 localization, the focus is on distinguishing foreground objects from background objects, as video recognition models heavily rely on foreground objects. The agent is trained to identify patches containing foreground objects using edgeboxes [34], which provide an objectness score for a patch. Edgeboxes calculate the edge response of each pixel and determine object boundaries using a structured edge detector. A higher overlap between the edgebox's output and the selected patch results in a higher reward.

$$r_{edgebox} = \frac{\sum_k w_b(s_k) \times u_k}{2 \times (w + h)^2} \quad (5)$$

Thus, the overall objectness reward for the entire video can be defined as:

$$r_1 = \sum_{i=1}^{r} i_{edgebox} \quad (6)$$

$w$ and $h$ represent the width and height of the patch respectively. The term $w_b(s_k)$ serves as a metric to quantify the correlation within the k-th groups of edges found within the chosen patch. The variable $u_k$ indicates the cumulative sum of the k-th edge groups within the selected patch.

### 5.2.4. Level-2 localization

With the agent identifying a high-level region of interest for the given video frames, we further divide the selected patch from the level-1 localization to get a finer estimate of the region of interest. This ensures the maintenance of the lower $l_p$ norm. For every frame that has been chosen by the temporal agent and the localization region extracted, we further divide the space into finer non-overlapping patches. The state and action are similar to the level-1 localization.

For the level-2 reward, our focus is on the specific region within the chosen patch that holds a higher amount of saliency information throughout the video sequence. This involves computing the motion saliency value within the selected region of spatiotemporal areas P (derived from the chosen patches in Level 1). The motion saliency is represented as a mapping $p \rightarrow R$, where p corresponds to a spatial partition of the frame's patch. To calculate the motion saliency, we employ the optical flow equation, utilizing the Lucas-Kanade method for estimation. Consequently, the Level 2 reward for the entire video is formulated as follows:

$$r_2 = \exp\left(\sum_{i=1}^{M} s(p_i)\right) \quad (7)$$

Here, $s(p_i)$ denotes the motion saliency value of the region and lies within the range [0, 1].

Finally, similar to the temporal agent, the reward $r_3$ is the final feedback from the victim models. It measures the impact of the added noise on a selected patch. If the selected patch is reasonable, the generated adversarial patch should have a strong impact on decreasing the confidence score of the initial prediction. This is represented as,

$$r_3 = exp(P(y|x'_t) - P(y|x)) \quad (8)$$

where $x$ represents the original video and $x'$ represents the perturbed video at time $t$.

### 5.3. Reverse Distortion Removal

Once the adversarial sample $x'$ has been generated, with the motivation to reduce distortion added to the adversarial sample, a distortion reversion process is performed. We iteratively attempt to remove distortions from the least sensitive region to which distortions were added while maintaining the misclassification and eventually reversing the distortions. This process is started once a misclassification has been This reduced the net distortions that were added to the sample by a significant margin.

---

**Algorithm 1** black-box video attack algorithm
**Input:** Clean video dataset; ground-truth label; Backbone CNN architecture:$f(\cdot)$; black-box video recognition model:$F(\cdot)$. Max iterations:$T$; learning rate:$\epsilon$.
**Output:** Trained spatial and temporal policy networks

1: Initialize parameters $\theta_t$ and $\theta_p$ for temporal policy network $\pi_t(\cdot)$ and spatial policy network $\pi_p(\cdot)$
2: **for** sample $x$ in the dataset with corresponding $y$ **do**
3:     Extract frame-level features with the backbone CNN architecture via $f\_feat_i = f(x)$, $i = 1, 2, \ldots, M$
4:     **while** $t < T$ **do**
5:        Compute key frames $A_t = \{a^*_i | i = 1, \ldots, M\}$
6:        Compute L-1 focus regions $A_{p1}^i = \{a^*_i | i = 1, \ldots, M\}$
7:        Compute L-2 focus regions $A_{p2} = \{a^*_i | i = 1, \ldots, M\}$ from the patch extracted from L-1.
8:        Generate adversarial video $X'_{t+1}$ via distortion type chosen to the selected frames and patches.
9:        Compute spatial reward $r^t_{spatial}$ and temporal reward $r^t_{temporal}$
10:       Update $\theta_t \leftarrow \theta_f + \epsilon \cdot \nabla_{\theta_t} J(\theta_t)$
11:       Update $\theta_p \leftarrow \theta_p + \epsilon \cdot \nabla_{\theta_p} J(\theta_p)$
12:       **if** $F(X'^{t+1}_{t+1}) = y$ **then**
13:          $X' \xleftarrow{t+1} X'_{t+1}$
14:          Break
15:       **end if**
16:     **end while**
17:     Reverse noise that was added, while retaining misclassification.
18: **end for**

---

### 5.4. RL Optimization

To facilitate this optimization process, we have specifically employed the Proximal Policy Optimization (PPO) algorithm [22]. PPO guides the adjustments within the policy networks, contributing to stable and effective policy updates. This results in enhanced convergence and learning within the actor-critic framework. The comprehensive algorithmic representation of the proposed attack approach is encapsulated in Algorithm 1.

## 6. Experiments and Results

### 6.1. Datasets

In our experiments, two popular and publicly available action recognition datasets are used, UCF-101 and HMDB-51 [9, 23]. The UCF-101 dataset contains 13,320 videos with 101 action classes and the HMDB-51 dataset contains 51 action classes with a total of 7000 videos. For all our experiments, we used 70 percent of the videos for training and 30 percent of the videos for validation. Samples that were recognized correctly were only considered for both training

| Threat models | Attack Methods | HMDB-51 | | | UCF-101 | | |
|---|---|---|---|---|---|---|---|
| | | MAP ↓ | QN ↓ | SR ↑ | MAP ↓ | QN ↓ | SR ↑ |
| TSM [12] | VBAD attack | 6.361 | 1818 | <u>92</u> | 6.023 | 2803 | 84 |
| | Heuristic attack | 5.043 | 10385 | 58 | 5.956 | 10657 | 40 |
| | Sparse attack | 3.334 | 6244 | 62 | 3.417 | 8529 | 58 |
| | M-S attack | 7.229 | 3911 | 90 | 7.237 | 5187 | 83 |
| | G-T attack | 5.919 | 3164 | <u>92</u> | 5.865 | 3782 | 88 |
| | RLSB attack | 5.323 | 5950 | <u>82</u> | 4.823 | 4898 | 87 |
| | AstFocus attack | 3.411 | 1529 | **100** | 3.355 | 1138 | **96** |
| | **Ours (GB)** | **0.835** | 921 | 83 | **0.735** | <u>863</u> | 85 |
| | **Ours (DP)** | 1.824 | **600** | 90 | 2.491 | **306** | <u>94</u> |
| | **Ours (GN)** | <u>1.016</u> | 802 | 88 | <u>1.469</u> | 865 | 84 |
| TSN [25] | VBAD attack | 5.873 | 2373 | 90 | 6.168 | 2450 | 84 |
| | Heuristic attack | 5.395 | 10146 | 58 | 5.265 | 9135 | 51 |
| | Sparse attack | 3.271 | 6765 | 74 | 3.131 | 6916 | 64 |
| | M-S attack | 7.275 | 3667 | 88 | 6.895 | 4744 | 78 |
| | G-T attack | 5.192 | 3392 | 88 | 5.472 | 3782 | 75 |
| | RLSB attack | 5.312 | 4217 | 92 | 5.238 | 3504 | 93 |
| | AstFocus attack | 3.520 | 2198 | <u>96</u> | 3.265 | 2015 | **99** |
| | **Ours (GB)** | **0.677** | <u>433</u> | <u>96</u> | **0.676** | 3243 | 91 |
| | **Ours (DP)** | 2.373 | **238** | **98** | 2.417 | **373** | <u>96</u> |
| | **Ours (GN)** | <u>1.987</u> | 948 | 94 | <u>1.380</u> | 886 | 94 |
| C3D [5] | VBAD attack | 6.743 | 4107 | 78 | 6.800 | 4890 | 75 |
| | Heuristic attack | 4.838 | 10534 | 42 | 6.295 | 14160 | 30 |
| | Sparse attack | 2.983 | 8545 | 46 | <u>3.009</u> | 9507 | 42 |
| | M-S attack | 7.035 | 6491 | 68 | 6.153 | 8132 | 62 |
| | G-T attack | 5.666 | 5082 | 84 | 5.877 | 7045 | 75 |
| | RLSB attack | 4.688 | 7279 | 62 | 5.326 | 6568 | 68 |
| | SVAL | 3.28 | - | 59 | 2.44 | 9402 | 63 |
| | AstFocus attack | 3.835 | 3628 | 92 | 4.015 | 4224 | <u>90</u> |
| | **Ours (GB)** | **0.835** | 8710 | 95 | **2.258** | 1783 | 90 |
| | **Ours (DP)** | 1.824 | <u>378</u> | <u>99</u> | 6.351 | 2602 | 81 |
| | **Ours (GN)** | <u>1.016</u> | **320** | **100** | 5.661 | 2829 | **96** |
| SF [4] | VBAD attack | 6.528 | 5442 | 72 | 6.302 | 4089 | 77 |
| | Heuristic attack | 5.875 | 9094 | 54 | 5.869 | 12776 | 34 |
| | Sparse attack | 3.228 | 8977 | 56 | 3.164 | 8642 | 58 |
| | M-S attack | 7.163 | 6553 | 74 | 7.086 | 5166 | 77 |
| | G-T attack | 6.242 | 5741 | 78 | 5.712 | 4334 | 84 |
| | RLSB attack | 5.680 | 4495 | 84 | 5.586 | 4563 | 85 |
| | AstFocus attack | 4.078 | 2295 | 96 | 4.286 | 1435 | <u>93</u> |
| | **Ours (GB)** | **0.654** | 1103 | 78 | **0.704** | 935 | 89 |
| | **Ours (DP)** | 2.085 | **203** | **99** | <u>1.659</u> | **419** | **94** |
| | **Ours (GN)** | <u>1.731</u> | 359 | 97 | 2.402 | <u>852</u> | 86 |

Table 1. Competitors SVAL(2022) [28] VBAD (2019) [8], Heuristic (2020)[30], Sparse (2022)[28], M-S (2020) [36]. G-T (2021) [11], RLSB (2021)[26], AstFocus [29] (2023) **bolded values** indicate the top-performing scores, <u>underlined values</u> correspond to the second-best scores. GB, DP, and GN stand for Gaussian Blur (GB), Dead Pixels (DP), and Gaussian Noise (GN), respectively.

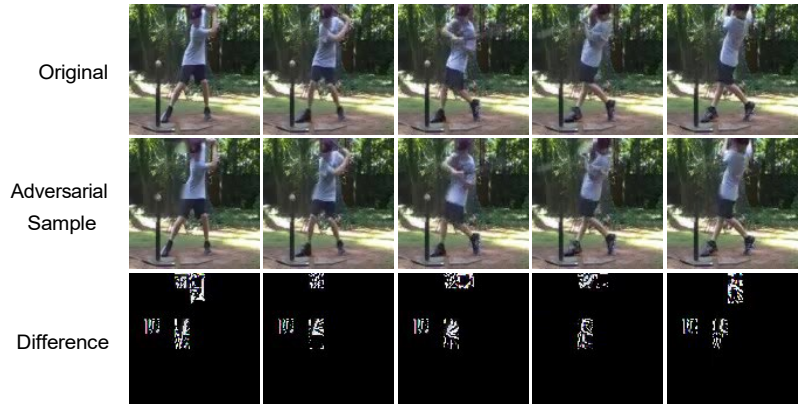

Figure 6. Sample output from the proposed method. The third row shows the difference between the original and the adversarial sample. Initially, the sample was predicted as **baseball**, which was misclassified to **shoot bow**. ∗ ∗ *zoomintogetaclearview*

and evaluation.

Table 2. Ablation study comparing different patch sizes

|  | HMDB-51 | | UCF-101 | |
| --- | --- | --- | --- | --- |
| 13x13 patch size | MAP ↓ | Query ↓ | MAP ↓ | Query ↓ |
| C3D | 1.22 | 3136 | 4.75 | 2404 |
| slowfast | 1.48 | 555 | 1.58 | 735 |
| tsm | 1.32 | 774 | 1.56 | 677 |
| TSN | 1.67 | 3206 | 1.48 | 1500 |
| 5x5 patch size | MAP | Query | MAP | Query |
| C3D | 0.39 | 851 | 1 | 1757 |
| SlowFast | 0.22 | 2361 | 0.17 | 397 |
| TSM | 0.34 | 1619 | 0.32 | 3858 |
| TSN | 0.35 | 1956 | 0.26 | 1947 |

Table 3. Ablation study showing the impact of different agents and rewards. The combined agent performs 21% better than the best individual agent measured by MAP.

| Agents with Rewards | MAP ↓ | Query Number ↓ |
| --- | --- | --- |
| Spatial Agent* | 0.975 | 136 |
| Temporal Agent* | 1.021 | 149 |
| All (two agents) | 0.81 | 110 |

*(corresponding rewards $r_1$, $r_2$ and $r_3$)

### 6.2. Video Recognition Models

For recognition models, four of the most popular and established architectures were considered. C3D, Temporal Segment Network (TSN), Temporal Shift Module (TSM), and SlowFast network [4, 5, 12, 25]. For all four video recognition models, we used pre-trained weights provided by the mmaction2 codebase framework [2].

### 6.3. Evaluation metrics

Success rate (SR): The success rate represents the percentage of adversarial videos that successfully fooled the victim models using the proposed method. A higher success rate indicates greater effectiveness of the method. $L_p$ norm: The $L_p$ norm, or p-norm, quantifies the difference between two samples, indicating how much one deviates from the other. This work focuses on $p = 1$ (Mean Absolute Perturbation, MAP), a common metric in adversarial attacks on images and videos. All competitors in Table 1 have reported MAP. Query Number (QN): The number of queries reflects the efficiency of video attack methods by denoting the number of query attempts needed to successfully fool the threat model with an adversarial video. This experiment sets an upper limit of 10,000 queries, and the average query number across test videos is reported.

Our training experiments were conducted on an Ubuntu machine with 8 Tesla V100S-PCIE-32GB GPUs and an Intel Xeon Gold 6246R CPU @ 3.40GHz with 16 cores. Training was distributed across multiple GPUs for configurations including all victim models and datasets. The maximum number of queries was set to 10,000.

### 6.4. Results and Discussion

We compare the proposed method with other recent state-of-the-art black-box attacks on two publicly available datasets on four action recognition models as victims. Sample output from the proposed method is shown in Figure 6. The table 1 presents an extensive experimentation of diverse threat models and their corresponding attack methods within the context of video action recognition, using the widely utilized HMDB-51 and UCF-101 datasets.

Upon close examination, it can be seen that variations of the proposed method (Gaussian Blur (GB), Dead Pixels (DP), Gaussian Noise (GN)) demonstrate a notable advantage over other established state-of-the-art methods. These variations showcase higher success rates, indicating their capacity to deceive the video recognition models. This can be observed for both HMDB-51 and UCF-101 datasets, underscoring their adaptability and generalizability. A similar scenario can be observed with all four types of victim models. While focusing on the number of queries, it can be observed that the values of our method are significantly lesser than the black-box attacks. We also observed that the magnitude of the distortion for Gaussian noise and the kernel size for Gaussian blur was maintained constant through the training. For Gaussian noise, a variance of 0.005 was a balance between the number of queries and the MAP with a grid search. Additionally, the choice of the number of hierarchical levels in localizing the spatial region was a computation vs performance tradeoff. We found that increasing the number of levels to more than three increased the complexity but with much lesser improvement in performance. This could be due to the nature of the dataset.

#### 6.4.1. Comparing different patch sizes and reward ablation

Table 2 compares 13x13 and 5x5 patch sizes where lower MAP and lower query numbers are desirable. For both the datasets larger patch size, higher the MAP, and lower the query numbers and vice versa. The choice between patch size depends on the specific priorities, whether the requirement is higher precision or query efficiency.

Table 3 compares the impact of individual agents with the combined agents. It shows that combining both the agents/rewards yields better results compared to the individual ones.

#### 6.4.2. Limitations

Typically, an exhaustive search can potentially yield a better performance at a prohibitive computational cost. However, our approach allows a power vs performance tradeoff and favors a cheaper deployment and carbon footprint reduction.

### 7. Conclusion

In this study, we introduce a novel adversarial attack method that uses a multi-agent RL strategy. Our approach involves the simultaneous identification of pivotal frames and key regions within videos to which flexible distortions can be added. This is achieved through a cooperative multi-agent RL framework, where one agent is dedicated to selecting crucial frames, while another agent focuses on identifying significant regions. These agents are collaboratively trained using shared rewards obtained from black-box threat models. Through iterative querying, the subset of essential frames and regions converges to a refined representation, considerably reducing the overall query count and $L_1$ distance compared to the original video. Our extensive experimentation across four prominent video recognition models and two widely used action recognition datasets attests to the efficacy and robustness of the proposed method. Furthermore, stable performance across different types of distortions considered suggests that the proposed framework is robust irrespective of the type of distortion.